\documentclass[10pt]{article} 
\usepackage[accepted]{tmlr}

\usepackage{amsmath,amsfonts,bm}

\def\eqref#1{equation~\ref{#1}}

\def\1{\bm{1}}

\DeclareMathAlphabet{\mathsfit}{\encodingdefault}{\sfdefault}{m}{sl}
\SetMathAlphabet{\mathsfit}{bold}{\encodingdefault}{\sfdefault}{bx}{n}

\usepackage{hyperref}       
\usepackage{url}            
\usepackage{booktabs}       
\usepackage{amsfonts}       
\usepackage{microtype}      
\usepackage{graphicx}
\usepackage[table]{xcolor}

\usepackage{url}            
\usepackage{amsfonts}       
\usepackage{nicefrac}       
\usepackage{array}
\usepackage{caption}
\usepackage{subcaption}
\usepackage{amssymb}
\usepackage{mathtools}

\usepackage{soul}
\usepackage{bbm}
\usepackage{tikz}
\usepackage{multirow}
\usepackage{wrapfig,lipsum}
\usepackage{colortbl}

\usepackage{algorithm}
\usepackage{algorithmicx}
\usepackage{algpseudocode}
\usepackage{amsmath}
\usepackage{amsthm}
\usepackage{pifont}  
\usepackage{makecell}
\usepackage[font=small,skip=4pt]{caption} 
\newcommand{\cmark}{{\color{green!60!black}\ding{51}}} 
\newcommand{\xmark}{{\color{red}\ding{55}}}             

\definecolor{grey}{rgb}{0.9, 0.9, 0.9}
\definecolor{amethyst}{rgb}{0.6, 0.4, 0.8}
\definecolor{darkmagenta}{rgb}{0.55, 0.0, 0.55}

\renewcommand{\th}{\boldsymbol{\theta}}

\renewcommand{\th}{\boldsymbol{\theta}}

\newcommand{\tm}[1]{\textcolor{black}{#1}} 
\newcommand{\tmh}[1]{\textcolor{black}{#1}} 

\title{Dual-Phase Continual Learning: Supervised Adaptation Meets Unsupervised Retention}

\author{\name Vaibhav Singh \email vaibhav.singh@mila.quebec \\
      \addr Department of Computer Science\\
      Mila, Concordia University
      \AND
      \name Rahaf Aljundi \email rahaf.al.jundi@toyota-europe.com \\
      \addr Toyota Motor Europe
      \AND
      \name Eugene Belilovsky \email eugene.belilovsky@concordia.ca\\
      \addr Mila, Concordia University
      }

\def\month{01}  
\def\year{2026} 
\def\openreview{\url{https://openreview.net/pdf?id=GFrHdXzZwo}} 

\begin{document}
\maketitle
\begin{abstract}
Foundational Vision-Language Models (VLMs) excel across diverse tasks, but adapting them to new domains without forgetting prior knowledge remains a critical challenge. Continual Learning (CL) addresses this challenge by enabling models to learn sequentially from new data while mitigating the forgetting of prior information, typically under supervised settings involving label shift. Nonetheless, abrupt distribution shifts can still cause substantial forgetting, potentially nullifying the benefits of supervised updates, especially when storing or replaying past data is infeasible. In this work, we propose leveraging unlabeled test-time data in an unsupervised manner to \textit{\textbf{reinforce prior task performance without requiring replay or stored examples}}. Unlike traditional Test-Time Adaptation (TTA), which primarily focuses on domain shift or corruption, our method improves performance on earlier tasks by exploiting representative test samples encountered during deployment. We introduce a simple Teacher-Student framework with gradient-based sparse parameter updates, and show that it effectively mitigates forgetting in class-incremental CL for VLMs, offering a memory-free alternative to episodic replay with strong empirical results.
\end{abstract}

\section{Introduction}
Foundation models in computer vision have shown impressive performance on various downstream tasks and domains, which renders them a key building block of various solutions, including generative vision language models \citep{li2022blip, chen2023minigpt, bommasani2021opportunities}. However, naively adapting these pre-trained models to distribution shifts or new tasks often leads to \textit{catastrophic forgetting} \citep{aMCCLOSKEY1989109}, where new learning sessions interfere with what a model has previously acquired. 

To address catastrophic forgetting, Continual Learning (CL) enables models to adapt to evolving data distributions over time. Key approaches include regularization-based methods \citep{ewc, lomanco, lwf, schwarz2018progress, singh2024wake}, external memory approaches \citep{lopez2017gradient, rolnick2019experience,  shin2017continual}, and dynamic model architecture techniques~\citep{dytox, NEURIPS2021_86a1fa88}. However, training from scratch often overlooks the rich representations learned by large pre-trained models. With the advent of foundation models, there is a growing interest in integrating CL with their representational power~\citep{ptm, clip, ridnik2021imagenet, caron2021emerging, dinov2}.
\begin{figure*}[t]
        \centering
\includegraphics[width=1\linewidth]{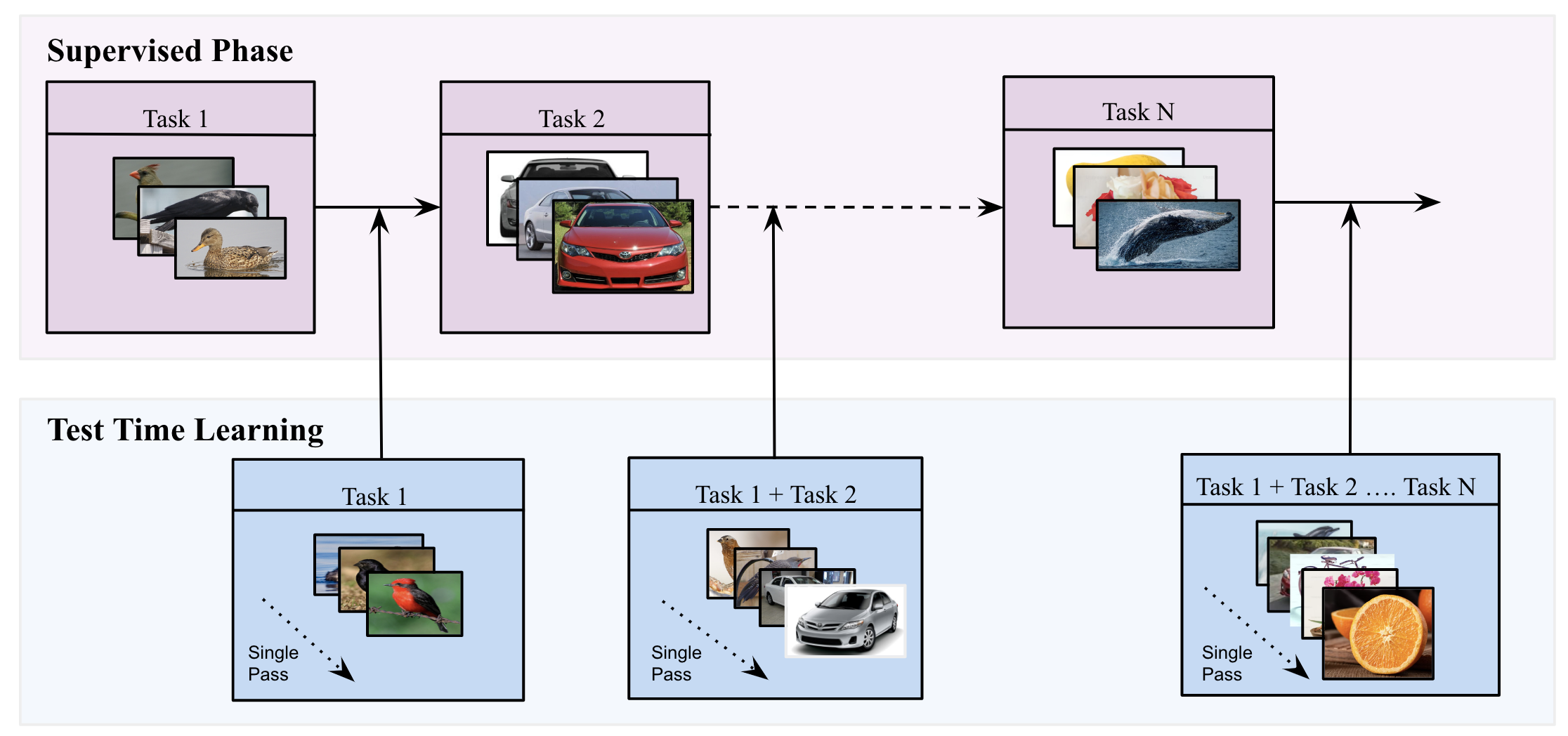} 
        \caption{\small An illustration of our proposed setting of \textbf{Continual Learning with Interleaved Test Time Learning.} After each supervised training session, the model is deployed to adapt in an unsupervised deployment phase, where it encounters data from both current and previously seen tasks. During this phase, the model adapts to the current task's classes while striving to preserve performance on earlier tasks, thereby mitigating forgetting.}
        \label{fig:cl_tta}
    \end{figure*}
    
Significant efforts have been made to improve foundational models' adaptability to new data streams \citep{learning_to_prompt, smith2023coda, zhou2023revisiting, slca, goyal2023finetune, wang2022dualprompt}, primarily through supervised training. However, this focus often leaves models static and prone to catastrophic forgetting \citep{wang2024comprehensive, prabhu2023computationally}. In contrast, the diverse unsupervised data encountered during inference presents an underexplored opportunity to mitigate forgetting without additional supervision. We consider a continual learning setup, especially the challenging scenario of class incremental learning (CIL), where supervised training phases are interleaved with unsupervised deployment, as shown in Figure \ref{fig:cl_tta}. To mitigate catastrophic forgetting, the model leverages unlabeled test-time data in an online manner—processing each sample once and discarding it, thereby reducing both privacy risks \citep{verwimp2023continual} and computational overhead.

To the best of our knowledge, we are among the early works that utilize test-time data for alleviating forgetting in continual learning, particularly through an interleaved test-time learning stage. We propose a novel Dual-Phase Continual Learning framework: \textbf{DoSAPP} (described in Section \ref{sec:method}) that uniquely combines supervised adaptation with unsupervised retention, all without relying on replay buffers or explicit task boundaries. Built upon the CLIP foundation model~\citep{clip}, which offers strong generalization and transfer~\citep{rasheed2023fine, pei2023clipping}, our approach introduces two distinct learning phases: supervised sessions enable efficient task-specific adaptation through sparse parameter updates while unsupervised sessions promote long-term stability by reinforcing previously acquired knowledge. Central to this design is the Teacher-Student \citep{tarvainen2017mean} framework governed by our novel \textit{dual-momentum} mechanism, which decouples the adaptation rates of the teacher and student to balance plasticity and stability. Additionally, cumulative mask consolidation across tasks ensures scalable memory retention without interference. This synergy enables robust continual learning in realistic, dynamically evolving environments.

Existing works such as Test-Time Adaptation (TTA) \citep{sun2020test, wang2020tent, zhang2022memo, niu2022efficient, online_tta} and Continual Test-Time Adaptation (CTTA) \citep{ctta, gong2022note, niu2022efficient, song2023ecotta, tian2024parameter, wang2024continual} similarly utilize unsupervised test-time data, but with a different focus: \textit{adapting models to unknown distribution shifts in data during deployment.} These methods consider previously unseen domain shifts and corruptions in the test-time data itself, aiming to perform well on this unsupervised data, while not degrading the performance of the model on past data. Further, these methods cannot address forgetting in a class incremental learning (CIL) setup. On the other hand, our setting, which interleaves supervised and unsupervised sessions, considers leveraging test-time data that does not require domain shift/corruption with respect to previously seen data. Instead, it aims to use the unsupervised data to combat forgetting in the CIL setting through the supervised sessions. Table~\ref{contrast_bet_ctta_dosapp} highlights the key differences between our proposed approach \textbf{DoSAPP} and existing CTTA methods.

 
Our contributions are as follows: 1) We propose a new setting for continual learning where test-time data can be leveraged for forgetting without explicit knowledge of task boundaries, especially in the challenging scenario of CIL. 2) We investigate different baselines for this setting. 3) We propose a novel approach that illustrates the utility of test-time data in supervised continual learning and the significant reduction in forgetting without relying on any external replay buffer.


\section{Related Work}
\label{sec:relatedwork}

\begin{table}[t!]
\centering
\caption{\small \textit{\textbf{Comparison of Continual Test-Time Adaptation (CTTA) methods with our approach, DoSAPP}}. Our work is fundamentally orthogonal to TTA and CTTA methods, which primarily tackle distribution shifts during test time. Moreover, existing CTTA approaches do not typically address forgetting of previous tasks (classes) upon the introduction of new ones. These methods are also incompatible with class incremental learning (CIL). In contrast, DoSAPP leverages test-time data to enhance CIL performance and reduce forgetting through unsupervised learning on test samples.}
\resizebox{\textwidth}{!}{%
\begin{tabular}{lccccc}
\toprule
\textbf{Method} &
\makecell{\textbf{Corrupted} \\ \textbf{Data}} &
\makecell{\textbf{Uncorrupted} \\ \textbf{Data}} &
\makecell{\textbf{Interleaved} \\ \textbf{Supervision}} &
\makecell{\textbf{Reduces} \\ \textbf{Forgetting}} &
\makecell{\textbf{Handles} \\ \textbf{New Classes (CIL)}} \\
\midrule
CoTTA (2022) \citep{ctta}  & \cmark & \xmark & \xmark & \xmark & \xmark \\
NOTE (2022) \citep{gong2022note}  & \cmark & \xmark & \xmark & \xmark & \xmark \\
EATA (2022) \citep{niu2022efficient} & \cmark & \xmark & \xmark & \cmark & \xmark \\
EcoTTA (2023) \citep{song2023ecotta} & \cmark & \xmark & \xmark & \cmark & \xmark \\
RMT (2023) \citep{dobler2023robust} & \cmark & \xmark & \xmark & \xmark & \xmark \\
PSMT (2024) \citep{tian2024parameter} & \cmark & \xmark & \xmark & \xmark & \xmark \\
DSS (2024) \citep{wang2024continual} & \cmark & \xmark & \xmark & \xmark & \xmark \\
\midrule
\textbf{DoSAPP (ours)} & \cmark & \cmark & \cmark & \cmark & \cmark \\
\bottomrule
\end{tabular}
}
\vspace{-10pt}
\label{contrast_bet_ctta_dosapp}
\end{table}
\textbf{Continual Learning from Pre-trained Models:} Continual learning with pre-trained models is gaining popularity due to the availability of powerful foundation models \citep{clip, dinov2, LLM}. Recent approaches \citep{koh2022online, boschini2022transfer} employ a Teacher-Student framework with knowledge distillation but rely on memory buffers to mitigate catastrophic forgetting, which can be memory-intensive \citep{zhou2022model, prabhu2023computationally}. Additionally, stored logits become outdated, requiring updates with task boundary information, which may not always be available in task-free CL. Further recent works such as \citep{chen2024adaptive} utilize OOD-based techniques to calibrate only the classifier layer of a pretrained model using test-time samples, aiming to improve performance in CIL setting. While our method also operates on the test stream, it differs fundamentally from \citep{chen2024adaptive}: their approach requires access to task boundaries, which may not be available at inference. In contrast, our method makes no such assumptions and remains applicable even when task boundaries are unknown.

To address these limitations, CLIP \citep{clip} presents an attractive alternative as it inherently avoids classification head issues and retains a broad feature space, making it well-suited for continual learning.  Inspired by SPU \citep{spu}, which preserves generic knowledge by modifying a sparse subset of parameters based on gradient scoring, we explore leveraging test data in continual learning. This opens avenues for self-supervised techniques to enhance feature representations while mitigating recency bias. 

\textbf{Test Time Adaptation (TTA):} TTA methods primarily focus on handling domain shifts and adapting to data corruption in test data. These approaches aim to improve model performance on the adapted test data, often leveraging techniques such as self-supervised learning \citep{online_tta}, batch normalization \citep{nado2020evaluating, vianna2024channel}, entropy minimization \citep{wang2020tent, niu2023towards}, pseudo labeling \citep{chen2022contrastive, shott}, and continually adapting to varying test-time distribution shifts \citep{ctta, gong2022note, niu2022efficient, song2023ecotta, tian2024parameter, wang2024continual}. In contrast, our approach leverages test-time data for a broader purpose: mitigating forgetting of past tasks. Unlike TTA methods, which leverage the test-time data to correct domain shifts, our method utilizes unsupervised test-time data from previous tasks to \textit{\textbf{retain past knowledge obtained in the supervised training sessions.}}

\section{Methodology}
\label{sec:method}
We propose a novel class incremental continual learning (CIL) setting using test-time data. As shown in Figure \ref{fig:cl_tta}, the model recovers lost knowledge from previously seen tasks after each supervised session, adapting to new classes while minimizing forgetting without relying on an external replay buffer. We consider a setting where supervised datasets $[{\mathcal{D}_1^{s}, \mathcal{D}_2^{s}, .....\mathcal{D}_T^{s}}]$ drawn from different distributions arrive incrementally at training sessions $t$ ranging from $0$ to $T$. Each session $t$ with $N_t$ instances includes dataset $\mathcal{D}_t^{s} = {(\mathbf{x}^t_i, y^t_i)}^{N_t}_{i=1}$ where an instance $\mathbf{x}_i^t \in \mathbb{R}^D$ belongs to class $y^t_i \in Y_t$, with disjoint label spaces $Y_t \cap Y_{t'} = \phi$ for $t\neq t'$. At session $t$, only the current dataset $\mathcal{D}^s_t$ is available for training the model $\mathcal{M}(\th)$. 

After training on $\mathcal{D}_t^{s}$, the model is deployed until $\mathcal{D}_{t+1}^{s}$ becomes available. During this phase, it encounters unsupervised test-time data $\mathcal{D}_t^{u}$, drawn from all previously seen tasks as shown in Figure~\ref{fig:cl_tta}. We leverage this data for online unsupervised adaptation to mitigate forgetting. Evaluation is performed on distinct test datasets $\mathcal{D}^e_t$ to ensure proper generalization assessment.

We further note that although supervised phases may permit multiple passes through the data until convergence, it would be impractical to collect unsupervised data in production and then perform adaptation on it. We thus restrict the unsupervised phase to be in the online setting \citep{online_tta, online_tta_2, online_learning}. This is especially important in cases where data privacy is a constraint, e.g., an assistant robot in a private smart home environment.


\subsection{\texorpdfstring{DoSAPP: \underline{Do}uble \underline{S}moothing via \underline{A}ffine \underline{P}rojected \underline{P}arameters}{DoSAPP: Double Smoothing via Affine Projected Parameters}}
\begin{figure*}[t]
 \centering
{\includegraphics[trim=10 20 10 10, clip, width=0.99\textwidth]{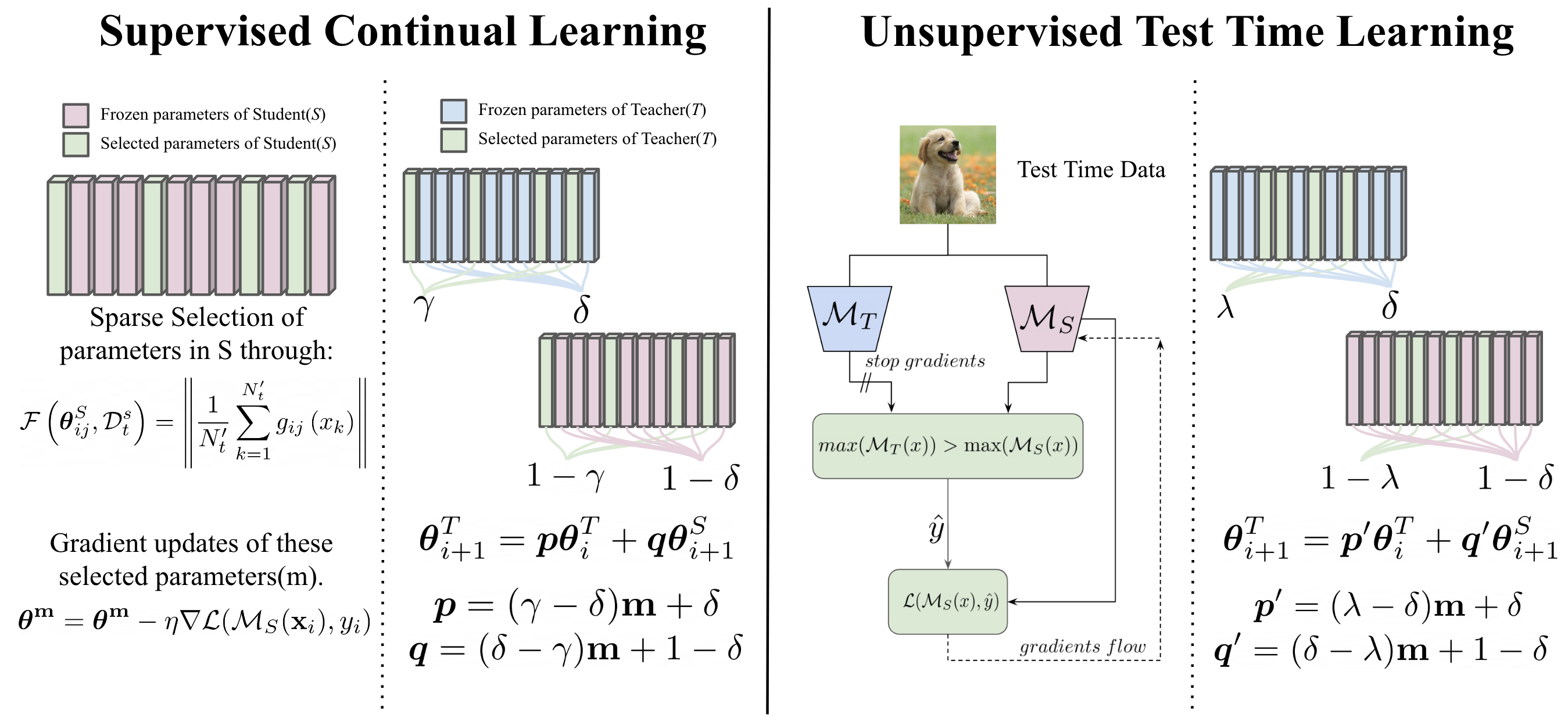}} 
\caption{ \small DoSAPP employs Teacher-Student ($\mathcal{M}_T,\; \mathcal{M}_S $) models respectively. In the Supervised Continual Learning phase, $\mathcal{M}_S$ performs sparse parameter selection using a gradient-based scoring function $\mathcal{F}$, followed by training on the selected parameters $\th^{\textbf{m}} \in \th^S$. After each update, $\mathcal{M}_T$ parameters $\th^T$ are updated through weighted exponential smoothing based on the affine projection of the boolean mask $\textbf{m}$, controlled by dual momentum terms $\delta$ and $\gamma$ for $\mathcal{M}_T$ and $\mathcal{M}_S$, respectively. In the unsupervised test-time learning phase, $\mathcal{M}_S$ adapts using ``pseudo-label'' derived from $\mathcal{M}_T$-$\mathcal{M}_S$ logits comparison. $\mathcal{M}_T$ then undergoes weighted smoothing again, with momentum terms $\delta$ and $\lambda$ for $\mathcal{M}_T$ and $\mathcal{M}_S$ (where $\gamma<\lambda<\delta$). This two-phase approach ensures generalization over previous knowledge while maintaining adaptability to new tasks.}
\label{fig:illustration}
\end{figure*} 

We propose a simple yet effective method for continual test-time learning, Double Smoothing via Affine Projected Parameters, aka DoSAPP. Our approach combines two key components:
\textbf{1) sparse and local updates:} to reduce forgetting, maintain generalization, by constraining adaptation to a small set of parameters, and \textbf{2) Teacher-Student framework} to promote stability in online updates and minimize forgetting.
In the continual test time learning, we can identify two distinct phases of learning, as outlined below.

\subsection*{Phase 1: Continual Learning Supervised Training with Sparse Selected Parameters}
\label{sup_phase}
Our goal is to rapidly acquire new knowledge while preserving generic knowledge during both training and testing. To achieve this, we update only a small subset (sparsity threshold $c=10\%$ of parameters. \tm{We choose $c=10\%$ for all main experiments because it offers a consistent efficiency–performance trade-off across tasks and ensures stable behavior during long horizon test time adaptation.} It has been demonstrated that updating only a small subset of relevant parameters in pre-trained models like CLIP can significantly reduce forgetting \citep{spu}. Complementary findings indicate that the MLP blocks in transformers act as key-value memory components, with the first MLP layer serving as a pattern detector \citep{geva2020transformer}. This implies that updating only the first MLP layer may suffice for retaining prior knowledge. Thus, we restrict updates to the first MLP layer of each transformer block in CLIP. From these candidates, we select the top-K (K=$c$) parameters, where $c$ is the sparsity threshold. \tm{This results in sparse, localized parameter updates, instead of broad model changes which helps preserve prior knowledge.} For the sake of simplicity, model parameters in the paper refer to these candidate parameters. 

Following \citep{spu}, we use the gradient magnitude of the loss with respect to the input data to assess parameter relevance, where a higher magnitude indicates a greater expected reduction in loss. We optimize the model $\mathcal{M}_{S}$ by first identifying the most relevant parameters from the candidate parameters of first MLP layer of each transformer block, $\th^{\textbf{m}}$ ($\th^{\textbf{m}} \in \th^S$), using:\begin{equation}
\label{scoring_spu}
\mathcal{F}\left (\th_{i j}^{S}, \mathcal{D}^s_t\right)=\left\|\frac{1}{N_t^{\prime}} \sum_{k=1}^{N_t^{\prime}} g_{i j}\left (x_k\right)\right\|,
\end{equation} where $g_{i j}\left (x_k\right)$ is the gradient of the CLIP loss $\mathcal{L} (\mathcal{M}_S, x_k, y_k)$ w.r.t. parameter $\th_{i j}^S$, computed per data point $(x_k, y_k) \in \mathcal{D}^s_t$. Iterating over the dataset once, we compute these scores and select the top-K ($K=c$) most relevant parameters based on a sparsity threshold $c=0.1$. This results in a binary mask $\textbf{m}$, freezing all but the selected parameters.

\begin{algorithm}[tb]
\caption{DoSAPP algorithm for continual and test time learning}
\label{alg:CPL}
\begin{algorithmic}[1]
\Require $\mathcal{M}_{S}(\boldsymbol{\theta}^S)$, CLIP loss: $\mathcal{L}(.,.,.)$, sparsity threshold $c$, learning rate $\eta$.

\State $\boldsymbol{\theta}^T \gets \boldsymbol{\theta}^S$ 
\For{$t$ in tasks}
    \State $\boldsymbol{\theta}^\mathbf{m} \leftarrow$ top-K (K=$c$) params from MLP layers of $\boldsymbol{\theta}^S$ based on $\mathcal{F}$ \Comment{\textcolor{blue}{Sparse Selection, Eq. \ref{scoring_spu}}}

    \For{($x_i, y_i$) in $\mathcal{D}^s_t$}
        \State $\boldsymbol{\theta}^\mathbf{m} \gets \boldsymbol{\theta}^\mathbf{m} - \eta \nabla \mathcal{L}(\mathcal{M}_S(x_i), y_i)$ \Comment{\textcolor{blue}{Take one SGD step}}

        \State $\boldsymbol{\theta}^{T}_{i+1} \gets p \boldsymbol{\theta}^{T}_i + q \boldsymbol{\theta}^{S}_{i+1}$ \Comment{\textcolor{blue}{Dual momentum for teacher EMA update, Eq \ref{eq:ema_dual}}}
    \EndFor

    \State Compute union of masks for all tasks seen so far $\mathbf{m}_u$ \Comment{\textcolor{blue}{Start of Unsupervised Phase}}
    \State Select $\mathbf{m}_u$ params in $\mathcal{M}_{S}$

    \For{$x_i$ in $\mathcal{D}^u_t$}
        \State $l_T \gets \max(\mathcal{M}_T(x_i), \text{dim}=1)$
        \State $l_S \gets \max(\mathcal{M}_S(x_i), \text{dim}=1)$
        \If{$l_T > l_S$}
            \State $\hat{y} \gets \arg\max(\mathcal{M}_T(x_i))$
        \Else
            \State $\hat{y} \gets \arg\max(\mathcal{M}_S(x_i))$
        \EndIf

        \State $\boldsymbol{\theta}^{\mathbf{m}_u} \gets \boldsymbol{\theta}^{\mathbf{m}_u} - \eta \nabla \mathcal{L}(\mathcal{M}_S(x_i), \hat{y})$ \Comment{\textcolor{blue}{Take one SGD step}}

        \State $\boldsymbol{\theta}^{T}_{i+1} \gets p' \boldsymbol{\theta}^{T}_i + q' \boldsymbol{\theta}^{S}_{i+1}$ \Comment{\textcolor{blue}{Dual momentum for teacher EMA update, Eq \ref{eq:ema_dual_ttl}}}
    \EndFor
\EndFor
\end{algorithmic}
\end{algorithm}

\subsubsection*{Teacher Student Framework}
To ensure stability later during online updates and reduce forgetting, we utilize a Teacher-Student framework~\citep{mt, koh2022online, boschini2022transfer, dobler2023robust, michael_momentum} where the student model is denoted by~$\mathcal{M}_{S} (\th^S)$ and the teacher model is denoted by~$\mathcal{M}_{T} (\th^T)$. 

During both supervised and unsupervised phases, the teacher model $\mathcal{M}_{T}$ parameters $\th^T$ are updated using the exponentially moving average (EMA) of the student model parameters~$\th^S$. Typically, in the Teacher-Student framework, all teacher parameters move toward the student parameters with a single smoothing parameter (momentum). However, we demonstrate that a single smoothing parameter is insufficient and leads to suboptimal performance, as shown in our ablation studies (Table~\ref{tab:ablation_component}) and in the Appendix (Table~\ref{tab1_clip_mom}). In our setting, most of the student model parameters (candidate parameters of the first MLP block) remain frozen, with only a small subset (sparsity threshold, $c=10\%$) being updated. We propose that the teacher model parameters corresponding to frozen candidate parameters of the student model should adapt at a different rate than those associated with active parameters. To address this, we introduce dual smoothing parameters (dual momentum), adjusting teacher parameter updates based on an affine transformation of the binary mask $\textbf{m}$. It should be noted that the non-candidate parameters (e.g., attention blocks) always remain frozen and are not updated at all.

\subsubsection*{Weighted exponential smoothing with dual momentum}
\label{ema_phase1}
After each gradient update step ($i$) for $\mathcal{M}_S$, parameters of $\mathcal{M}_T$ are updated by EMA of the student model parameters. Typically, EMA is governed by 
\begin{equation}\label{eq:ema_typ}
   \th^{T}_{i+1} = \delta \th^{T}_i + (1-\delta) \th^{S}_{i+1} ,
\end{equation} where $\delta$ is the smoothing parameter. Further, it has been shown in \citep{mt, dinov2, koh2022online} that setting $\delta$ to a high value (e.g., 0.998) maintains a stable teacher model that can be considered as a strong reference for past tasks $\{0,\dots, t-1\}$. But updating the teacher model with a single smoothing parameter in cases where parameters are masked creates dissonance and increases forgetting because all the parameters are updated with equal importance, disregarding those parameters which are selected by the gradient scoring function (where $[\textbf{m}_{ij}=1]$). To account for masking, we modify Eq. \ref{eq:ema_typ} as 
\begin{equation}\label{eq:ema_dual}
   \th^{T}_{i+1} = \boldsymbol{p} \th^{T}_i + \boldsymbol{q} \th^{S}_{i+1} ,
\end{equation}

where $\boldsymbol{p}$ and $\boldsymbol{q}$ denote the smoothing parameters for the teacher and student model, respectively, and can be computed as 
\begin{gather}
    \boldsymbol{p} = (\gamma - \delta)\mathbf{m} + \delta, \notag \\
    \boldsymbol{q} = (\delta - \gamma)\mathbf{m} + 1 - \delta
    \label{eq:P_Q_ema3}
\end{gather}
where $\gamma < \delta$. This implies that the selected parameters of the teacher model ([$\textbf{m}_{ij}=1$]) move slightly faster towards the student model as compared to the frozen candidate parameters (where  [$\textbf{m}_{ij}=0$]). As such,  parameters where  [$\textbf{m}_{ij}=0$] will move at a slow rate of $\delta$, and unmasked parameters would be updated with  $\gamma$. When $\gamma=\delta$, the weighted scheme becomes EMA with a single smoothing parameter. A detailed proof is given in Appendix \ref{proof_dual_mom}.

\subsection*{Phase 2: Unsupervised Test Time Learning (TTL)}
\label{unsup_phase}
After supervised training is completed, both $\mathcal{M}_{T}$ and $\mathcal{M}_{S}$ are deployed for Test Time Learning (TTL). We consider teacher ($\mathcal{M}_T$) and student ($\mathcal{M}_S$) models as two experts on different data distributions, the $\mathcal{M}_S$ on the most recent and the $\mathcal{M}_T$ on previous sessions distributions.

We take inspiration from Out-of-Distribution (OOD) literature~\citep{ood_softmax}, where a predictor assigns high scores to In-Distribution (ID) samples. Recent work~\citep{maxlogit} shows that using the unnormalized maximum logit as an ID score is more robust than softmax probability, which can overconfidently classify unknown samples~\citep{ood_confidence}. For CLIP, this logit corresponds to the cosine similarity between the image batch and text features. Following~\citep{maxlogit}, we use the maximum logit value of each expert as an ID score and select for each test sample the expert with the highest ID score, indicating that the sample is likely to be better represented by said expert. We then accept the pseudo-label of the selected expert. Formally, the pseudo label can be calculated as follows:
\begin{equation}
    \hat{y}= \begin{cases}\hat{y_T} & \text { if } l_T \geqslant l_S \; \\ \hat{y_S} & \text { otherwise}\end{cases} , 
\end{equation} where  $\hat{y}$ is the accepted pseudo label and $l_T=\max (\mathcal{M}_T (\textbf{x}))$ and $l_S=\max (\mathcal{M}_S (\textbf{x}))$ are the maximum logit score for teacher and student model respectively, and similarly $\hat{y_T}=\arg\max (\mathcal{M}_T (\textbf{x}))$ and $\hat{y_S}=\arg\max (\mathcal{M}_S (\textbf{x}))$ are the pseudo labels by teacher and student models respectively. During test-time training, the student model $\mathcal{M}_{S}$ is updated by minimizing CLIP contrastive loss given a pseudo label $\hat{y}$. In realistic settings, multiple iterations on test data are often not always possible, for example, in a streaming data pipeline. We too mimic this setting, where the entire data is processed only once during the TTL phase.

Similar to the above-mentioned supervised phase, we also apply sparse local updates to $\mathcal{M}_{S}$.  However, the estimation of masks based on the online data might be noisy and largely reduce the efficiency, as gradients of all parameters must be estimated for each mini-batch of test samples. To overcome this, and following the assumption that test data are drawn from the distributions of all previous tasks, we leverage the masks estimated for previous tasks\footnote{Parameters not relevant to the current stream of tasks will remain frozen, maintaining models' unrelated generic knowledge.}.
We accumulate a union of the binary masks ($\textbf{m}_u$) over all the previously seen tasks $t$ such that $\textbf{m}_u = \textbf{m}_1 \cup \textbf{m}_2 \cup ......\textbf{m}_t$. To maintain the same sparsity level ($c=0.1$) of performed updates, we further select the same top-K (K=$c$) most relevant parameters from these new masked $\textbf{m}_u$ parameters based on their previously computed gradient scores. 

Finally, $\mathcal{M}_{T} (\th^T)$ is updated using the same dual momentum scheme, but with different smoothing vectors $\boldsymbol{p}^{\prime}, \boldsymbol{q}^{\prime}$ as: \begin{equation}\label{eq:ema_dual_ttl}
   \th^{T}_{i+1} = \boldsymbol{p}^{\prime} \th^{T}_i + \boldsymbol{q}^{\prime} \th^{S}_{i+1} ,
\end{equation} where $\boldsymbol{p}^{\prime} = (\lambda-\delta)\textbf{m} + \delta$ and $\boldsymbol{q}^{\prime} = (\delta-\lambda)\textbf{m} + 1-\delta$. In the TTL phase, the momentum parameter $\lambda$ is kept such that $\gamma<\lambda<\delta$. Similar to phase 1, the selected parameters $[\textbf{m}_u=1]$ of Teacher Model ($\th^T$)  move slightly faster towards the student model $\th^S$ as compared to frozen candidate parameters $[\textbf{m}_u=0]$, ensuring stability in case of frequent and possibly noisy online updates. We discuss the effect of different momentum values in \ref{eff_momentum}.
The algorithm can be fully understood as given in algorithm block \ref{alg:CPL}.

\section{Experiments}\label{sec:experiments}

\subsection{Setup}
\textbf{Architecture:} We apply DoSAPP to vision-language classification tasks, given their relatively robust knowledge measurement in such tasks. To ensure consistency across experiments, CLIP-ViT-B/16 \citep{clip} is used as the backbone in DoSAPP as well as in all the baselines. We report the accuracies recorded by the Teacher model. We refer to \citep{spu} for hyperparameter selection other than dual momentum, which are given in Appendix \ref{hparams}.  

\textbf{Datasets:} We consider five different vision datasets, three fine-grained (\textit{Aircraft} \citep{aircraft}, \textit{CUB} \citep{cub}, \textit{Stanford Cars}~\citep{cars}, one coarse dataset (\textit{CIFAR100}~\citep{cifar10}) and one out-of-distribution dataset (\textit{GTSRB} \citep{gstrb}). These datasets are chosen primarily based on their initially low zero-shot performance with CLIP pre-trained models. To form the
continual learning sequences, we split each dataset into 10 subsets with disjoint classes, composing 10 tasks. For all the datasets, the training data is used in the supervised learning phase. The test data is divided into two disjoint splits, $\mathcal{D}^u$ and $\mathcal{D}^e$, where $\mathcal{D}^u$ is used for unsupervised test-time learning and $\mathcal{D}^e$ is reserved for evaluation. This separation ensures a fair assessment of the method’s generalization performance.

\textbf{Evaluation Metrics:} After each supervised session $t_i$ and the following test-time adaptation session, we evaluate the model's test performance on holdout datasets from all $T$ tasks. To do this, we construct the matrix $R \in \mathbb{R}^{T\times T}$, where $R_{i,j}$ is the test classification accuracy of the model on task $t_j$ after observing the last sample from task $t_i$. Thus, we compute \textbf{Average Accuracy} (Acc. $= \frac{1}{T} \sum_{i=1}^T R_{T, i}$), \textbf{Average Forgetting} (F. $ = \frac{1}{T-1} \sum_{j=1}^{T-1} f_{j,T}, \; f_{j,T} = \max_{i \in \{1, \ldots, T-1\}} (a_{i,j} - a_{T,j})$)~\citep{lopez2017gradient, wang2024comprehensive, spu}. These metrics allow us to assess how well a continual learner solves a classification problem while overcoming forgetting.

\looseness=-1\textbf{Baselines:} We comprehensively compare our method against various baselines. Firstly, we evaluate our approach against the best fine-tuning method of CLIP, FLYP \citep{goyal2023finetune}. We further integrate FLYP classical continual learning components to evaluate their performance on the CLIP backbone, including ER \citep{rolnick2019experience}, weight regularization method, MAS \citep{mas}, and functional regularization methods LwF \citep{lwf} and PRD \citep{prd}. We combine these functional regularization methods with a replay buffer (ER). We further consider
the latest pre-trained model-based continual learning techniques L2P \citep{learning_to_prompt}, DualPrompt \citep{wang2022dualprompt}, and SLCA \citep{slca}. Finally, we compare to recent methods that target knowledge retention of foundation models ZSCL \citep{zheng2023preventing}, SparseCL \citep{wang2022sparcl}, and SPU \citep{spu}.

\section{Results}
\label{results}
\subsection{Comparison with CL Methods}
\label{cl:comparison}
We compare our method (DoSAPP) with recent and diverse approaches in the challenging scenario of class incremental learning (CIL), as shown in Table \ref{tab:clip_results}. DoSAPP achieves the largest improvement over simple fine-tuning (FLYP)~\citep{goyal2023finetune}. Compared to MAS \citep{mas}, DoSAPP increases performance by at least 16\% on Cifar100, highlighting that not all parameters require updating. DoSAPP also outperforms prompt-based methods L2P \citep{learning_to_prompt} and DualPrompt \citep{wang2022dualprompt} by 12\% and 17\% respectively on Cifar100. Even against dynamic network approaches like SLCA \citep{slca}, SparceCL \citep{wang2022sparcl}, ZSCL \citep{zheng2023preventing}, and SPU \citep{spu}, DoSAPP achieves state-of-the-art or comparable results across all datasets. This demonstrates that test-time data can enhance transferability and preserve learned knowledge. Notably, DoSAPP achieves strong performance without a replay buffer, and when provided a small buffer of just 20 samples per task, it significantly outperforms methods requiring large buffers.
\begin{table*}[t!]
\caption{\small Acc. (Average Accuracy, $\uparrow$)  and F. (Forgetting, $\downarrow$) of different methods all using CLIP ViT-B/16 backbone with trainable vision and text encoders, without any Replay Buffer in CIL scenario. DoSAPP can achieve positive backward transfer: forgetting is negative on the Cars data (without ER).}
    \resizebox{\textwidth}{!}{
    \setlength{\tabcolsep}{3pt}
    \begin{tabular}{l cc cc cc cc cc}
    \toprule
     \multirow{2}{*}{Method} & \multicolumn{2}{c|}{Aircraft} & \multicolumn{2}{c|}{Cars}  & \multicolumn{2}{c}{CIFAR100} & \multicolumn{2}{c|}{CUB}  & \multicolumn{2}{c}{GTSRB}  \\
     & Acc. ($\uparrow$) & F. ($\downarrow$) & Acc. ($\uparrow$) & F. ($\downarrow$) & Acc. ($\uparrow$) & F. ($\downarrow$) & Acc. ($\uparrow$) & F. ($\downarrow$) & Acc. ($\uparrow$) & F. ($\downarrow$)\\
    \midrule
    CLIP-Zeroshot \citep{clip} &  24.45 & - & 64.63 & - & 68.25 & - & 55.13 & - & 43.38 & -  \\ 
    \toprule 
    FLYP (\small{fine-tuning}) \citep{goyal2023finetune} & 18.63  & 39.93 & 51.64 & 25.65 & 46.26 & 37.78 & 45.74  & 26.62 & 21.76  & 55.48 \\

     
    MAS \citep{mas} & 33.69  & 27.50  & 69.43 & 9.18 & 63.88 & 21.16 &  61.72 & 12.05 & 42.04 & 25.38  \\

    ZSCL \citep{zheng2023preventing} & 30.96 & 15.65 & 67.79 & 8.27 & \textbf{80.50} & \textbf{1.05} & 61.09 & 7.69 & 62.92 & 13.54\\
 
    SPU \citep{spu} & 30.94 & 28.36 & 69.41 & 16.91 & 58.80 & 26.37 & 62.31 & 7.2 & 43.06 & 19.16 \\ 


    \midrule
    \textbf{DoSAPP} & \bf{39.14} & \bf{12.55} & 
             \bf{74.87} & \bf{-0.74}  &
             79.16 & 7.73 & 
             \bf{68.17} & \bf{2.15} & 
             \bf{72.33} & \bf{1.02} \\
           
           
           

    \midrule
    \midrule
    ER methods (ER=1000)\\
    \midrule
  \rowcolor{grey} FLYP (\small{fine-tuning}) + ER \citep{er} & 41.42 & 31.38 & 69.08 & 16.42 & 82.86 & 3.41 & 64.07 & 17.72 & \textbf{96.28} & \textbf{-7.48 }\\
    
   \rowcolor{grey} LWF + ER \citep{lwf} & 36.08 & 18.12 & 72.56 & 4.04 & 74.32 & 8.16 & 65.11 & 5.90 & 53.56 & 11.86 \\
    
   \rowcolor{grey} PRD + ER \citep{prd} & 37.11 & 17.35 & 74.08 & \textbf{3.75} & 79.66 & 3.10 & 65.92 & 6.55 & 63.00 & 12.44 \\

   \rowcolor{grey}  L2P + ER \citep{learning_to_prompt} & 32.20 & 21.73 & 67.04 & 11.22 & 67.71 & 18.81 & 64.04 & 6.82 & 75.45 & 2.68\\ 
    
    \rowcolor{grey} DualPrompt + ER \citep{wang2022dualprompt} & 26.61 & 17.20 & 63.30 & 18.67 & 61.72 & 19.87 & 64.38 & 12.94 & 69.65 & 8.43 \\

     \rowcolor{grey} SparseCL + ER \citep{wang2022sparcl} & 31.95 & 19.77 & 71.57  & 5.38 & 69.35 & 15.23 & 62.50 & 9.66 & 48.99 & 24.91 \\ 

     \rowcolor{grey} SLCA + ER \citep{slca} & 29.40 & 11.45 & 62.65 & 4.42 & 70.03 & 0.19  & 53.87 & 7.75 & 46.01 & 0.83 \\
    
  
  \rowcolor{grey} SPU + ER & 42.89 & 15.55 & 73.69 & 5.84 & 79.65 & 7.36 & 71.92 & 4.67 & 87.64 & 2.18 \\

  \midrule

    \rowcolor{grey} \textbf{DoSAPP + ER=200} & \textbf{47.32} & \textbf{8.10} & \textbf{79.17} & 3.92  & \textbf{88.41} & \textbf{-1.96} & \textbf{74.39} & \textbf{2.77} & 83.67 & 1.92 \\


    \bottomrule
    \end{tabular}
    }
    \label{tab:clip_results}
    \vspace{1pt}
\end{table*}

\begin{table*}[t!]
\vspace{12pt}
    \caption{\small Acc. (Average Accuracy, $\uparrow$) and F. (Forgetting, $\downarrow$) in evaluating DoSAPP on imbalanced test data (referred as imb. in the table below), demonstrating its effectiveness in mitigating forgetting while maintaining high accuracy. }
     \centering
     \resizebox{0.9\textwidth}{!}{  
    \setlength{\tabcolsep}{2.5pt}
    \begin{tabular}{l cc cc cc cc cc}
    \toprule
     \multirow{2}{*}{Method} & \multicolumn{2}{c|}{Aircraft} & \multicolumn{2}{c|}{Cars}  & \multicolumn{2}{c}{CIFAR100} & \multicolumn{2}{c|}{CUB}  & \multicolumn{2}{c}{GTSRB}  \\
     & Acc. ($\uparrow$) & F. ($\downarrow$) & Acc. ($\uparrow$) & F. ($\downarrow$) & Acc. ($\uparrow$) & F. ($\downarrow$) & Acc. ($\uparrow$) & F. ($\downarrow$) & Acc. ($\uparrow$) & F. ($\downarrow$)\\
    \midrule
    FLYP (finetune) &	16.20 & 42.85 & 47.92 & 29.40 &	43.30 & 40.12	& 41.63 & 29.61	& 18.52 & 58.70 \\
ZSCL &	26.89 & 23.80 &	64.58 & 10.85 &	62.10 & 12.67 &	57.77 & 10.54 & 58.45 & 17.91 \\
SLCA &	27.44 & 19.23 &	56.83 &  14.17 & 59.92 & 18.01 &	48.66  & 10.82 & 39.67 & 21.56 \\
    SPU &  30.94 & 28.36	& 69.41 & 16.91&	58.80&  26.37&	62.31 & 7.2	&43.06 & 19.16  \\ 
   \textbf{DoSAPP}  & \textbf{35.99} & \textbf{15.26} & \textbf{72.68} & \textbf{6.38} & \textbf{75.70} & \textbf{9.81} & \textbf{64.84} & \textbf{3.73} & \textbf{68.17} &\textbf{ 5.63} \\
    \bottomrule 
    \end{tabular}
    }
    \label{tab:imbalanced}
\end{table*}

\textbf{Imbalanced test data:} We evaluate DoSAPP under a more realistic and challenging setting where $D_u$ (unsupervised data) is class-imbalanced as shown in Table~\ref{tab:imbalanced}. Each task is sampled from a symmetric Dirichlet distribution with a concentration parameter equal to the task length, leading to severe class imbalance, and at times, the complete absence of certain classes. This setup mirrors real-world scenarios where test-time data is often skewed. For fair evaluation, we retain a balanced evaluation set. Remarkably, despite the imbalance, DoSAPP still yields significant gains over purely supervised CL methods. While its performance is lower than in the balanced setting, this is expected: the model naturally adapts more to frequently observed classes during TTL. In practice, such behavior is desirable, as performance on the test distribution, rather than on rarely observed classes, is often the priority at deployment. 


\subsection{Comparison with TTA+CL Methods}
\label{ttl:comparison}
We highlight the key innovation of our approach: \textit{leveraging unsupervised test-time data - readily available in production, to enhance continual learning}. In contrast, most existing CL methods are not designed to incorporate such data, limiting their adaptability in this setting. Here, we investigate whether approaches that leverage unsupervised data in an online setting are capable of fully benefiting from test-time data in our setting. To examine this, we combine the best-performing CL method, SPU, with a simple self-training mechanism (SPU + $D^u$), where pseudo-labels are assigned based on the model’s max logit output. We also integrate RMT \citep{dobler2023robust}, with SPU~\citep{spu} and SparsCL~\citep{wang2022sparcl}. Although more recent CTTA methods like DSS~\citep{wang2024continual} and PSMT~\citep{tian2024parameter} have been proposed, we select RMT due to its consistently lower mean classification error across all the benchmark corruption datasets, making it a stronger baseline. We find that our proposed method, DoSAPP, consistently outperforms all variants as shown in Table~\ref{tab:spu_ttl}. This highlights a critical limitation: \textit{Continual Test Time Adaptation (CTTA) methods, when combined with ongoing supervised learning, suffer from substantial forgetting due to their inability to adapt across a long sequence of shifting tasks}. DoSAPP addresses this with a principled use of dual momentum over masked parameters. Furthermore, RMT performs worse than basic pseudo-labeling in nearly all cases, confirming that TTA methods are ill-suited for continual learning with expanding task distributions. We also demonstrate in Appendix~\ref{dep_on_test_data} and~\ref{size_tta} that DoSAPP remains robust to noise in test-time data and scales effectively with varying amounts of unlabeled input.
\begin{table*}[t!]
  \caption{\small Acc. (Average Accuracy, $\uparrow$)  and F. (Forgetting, $\downarrow$) comparing DoSAPP with CIL methods like SPU \citep{spu}, SparsCL \citep{wang2022sparcl} integrated with one of the most recent CTTA method: RMT \citep{dobler2023robust}. It can be observed that fusing the typical CTTA method in the CIL pipeline exacerbates the catastrophic forgetting. DoSAPP, on the other hand, outperforms all of them by a significant margin on all the datasets. }
    \centering
     \resizebox{0.9\textwidth}{!}{
    \setlength{\tabcolsep}{2.5pt}
    \begin{tabular}{p{0.17\linewidth} cc cc cc cc cc}
    \toprule
     \multirow{2}{*}{Method} & \multicolumn{2}{c|}{Aircraft} & \multicolumn{2}{c|}{Cars}  & \multicolumn{2}{c}{CIFAR100} & \multicolumn{2}{c|}{CUB}  & \multicolumn{2}{c}{GTSRB}  \\
     & Acc. ($\uparrow$) & F. ($\downarrow$) & Acc. ($\uparrow$) & F. ($\downarrow$) & Acc. ($\uparrow$) & F. ($\downarrow$) & Acc. ($\uparrow$) & F. ($\downarrow$) & Acc. ($\uparrow$) & F. ($\downarrow$)\\
    \midrule
    SPU &  30.94 & 28.36	&69.41 & 16.91&	58.80&  26.37&	62.31 & 7.2	&43.06 & 19.16  \\ 
    
   SPU + $D^u$ &  27.72 & 24.86 &	68.91 & 7.34 &	74.09 & 10.43&	61.21 & 4.01&	60.17 & 6.94  \\

   SparsCL + RMT &  27.11 & 16.29 & 69.81 & 17.22 & 70.82 & 12.25 & 60.03 & 10.58 & 51.98 & 11.40 \\

   SPU + RMT & 29.33 & 15.10 & 62.32 & 21.95 &	63.06 & 23.28 &	63.87 & 6.34 &	54.13 & 17.56  \\
    
   \textbf{DoSAPP} & \textbf{39.14} & \textbf{12.55} & \textbf{74.87} & \textbf{-0.74} & \textbf{79.16} & \textbf{7.73} & \textbf{68.17} & \textbf{2.15} & \textbf{72.33} & \textbf{1.02} \\
    \bottomrule 
    \end{tabular}
    }
    \vspace{-1cm}
    \label{tab:spu_ttl}

\end{table*}


\subsection{Class Incremental Long Sequence scenario with domain shift}
\label{long:seq}
We consider a long sequence of tasks trained in a class-incremental manner. For these experiments, we combined 10 tasks from Aircraft \citep{aircraft} and 10 from Cars \citep{cars}, introducing a domain shift after the first 10 tasks. Table \ref{tab:clip_long_seq} shows that our method, DoSAPP, outperforms SPU and Finetuning. Unlike SPU and Finetuning, which exhibit recency bias, DoSAPP achieves better overall and first-task accuracy, with only a $3.8\%$ drop in current task accuracy (CTA). This demonstrates its ability to retain earlier knowledge while adapting effectively to new tasks. We also report per task forgetting for Finetuning, SPU, and DoSAPP highlighting the effectiveness of our proposed method in \autoref{appn_forgetting}.

\begin{table*}[t!]
 \caption{\small Average Accuracy (Avg Acc.), First Task Accuracy (FTA), Current Task Accuracy (CTA), and Forgetting (F.) measured for a long sequence tasks made by the concatenation of the Aircraft \citep{aircraft} and Cars \citep{cars} datasets.}
    \centering
    \resizebox{0.6\textwidth}{!}{
    \setlength{\tabcolsep}{6pt}
    \begin{tabular}{lcccc}
    \toprule
    Method (CLIP) & Avg Acc. ($\uparrow$) & FTA ($\uparrow$) & CTA ($\uparrow$) & F. ($\downarrow$) \\
    \midrule
    Finetuning (no TTL) & 35.24 
    & 5.90 
    & 
    \textbf{75.44 }
    & 
    16.87 
    \\
    SPU & 39.62 
    & 24.31 
    & 74.94 
    & 7.32 
    \\
    \textbf{DoSAPP}  & \textbf{45.01} 
    & \textbf{30.63} 
    & {71.13} 
    & \textbf{2.34} 
    \\
    \bottomrule
    \end{tabular}
    }
    \label{tab:clip_long_seq}
\end{table*}
\vspace{0.1cm}

\begin{table*}[t!]
\vspace{12pt}
\caption{\small Acc. (Average Accuracy, $\uparrow$) and F. (Forgetting, $\downarrow$) when different components of DoSAPP are incrementally added to the Teacher-Student framework referred to as A1. A2 denotes the sparse parameter selection added to A1. EMA ($\delta$) represents single momentum updates, while EMA ($\delta$, $\gamma$) refers to dual momentum updates. $\textbf{m}_u$ denotes the union of mask technique described in section \ref{unsup_phase}.}
    \centering
   \resizebox{\textwidth}{!}{
    \setlength{\tabcolsep}{2pt}
    \begin{tabular}{llcc cc cc cc cc}
    \toprule
      \multirow{2}{*}{ID} & \multirow{2}{*}{Description} & \multicolumn{2}{c|}{Aircraft} & \multicolumn{2}{c|}{Cars} & \multicolumn{2}{c|}{CIFAR100} & \multicolumn{2}{c|}{CUB} & \multicolumn{2}{c}{GTSRB} \\
     & & Acc. ($\uparrow$) & F. ($\downarrow$) & Acc. ($\uparrow$) & F. ($\downarrow$) & Acc. ($\uparrow$) & F. ($\downarrow$) & Acc. ($\uparrow$) & F. ($\downarrow$) & Acc. ($\uparrow$) & F. ($\downarrow$)\\
    \midrule
    A1 & Teacher-Student (EMA ($\delta$)) & 30.12 & 13.50 & 67.72 & 3.66 & 77.82 & \textbf{5.17} & 62.67 & 4.11 & 53.57 & 5.38 \\
    A2 & A1 + sparse params & 34.16 & 18.61 & 69.42 & 3.41 & 71.93 & 8.24 & 66.32 & 3.98 & 55.32 & 5.81 \\
    A3 & A2 + $\textbf{m}_u$ & 31.79 & 10.42 & 70.99 & 3.64 & 72.66 & 8.86 & 66.98 & 3.17 & 61.54 & 4.01 \\
    A4 & A2 + EMA ($\delta$, $\gamma$) & 35.49 & 11.53 & 72.14 & 3.58 & 75.93 & 8.02 & 67.28 & 3.41 & 64.15 & 3.20 \\
   
    \textbf{DoSAPP*} & \bf{A4 + $\textbf{m}_u$} & \textbf{39.14} & \textbf{12.55} & \textbf{74.87} & \textbf{-0.74} & \textbf{79.16} & 7.73 & \textbf{68.17} & \textbf{2.15} & \textbf{72.33} & \textbf{1.02} \\
    \bottomrule
    \end{tabular}
    } 
    \label{tab:ablation_component}
\end{table*}

\section{Ablation Study}
\label{ablation}
In this section, we quantitatively analyze the effect of different components of our proposed method, DoSAPP. We evaluate the effects of each component incrementally, as seen in Table~\ref{tab:ablation_component}. We begin with a baseline Teacher-Student setup using a single momentum (\textbf{A1}), and compare it with a variant that applies localized sparse updates to the first MLP layer of each transformer block (\textbf{A2}). This yields performance gains in 4 out of 5 datasets. Next, we introduce the union of supervised task masks ($\textbf{m}_u$) during the TTL phase (\textbf{A3}), which leads to marginal improvements across most datasets, except for Aircraft—where applying a single momentum to both masked and unmasked parameters causes a mismatch and accuracy drop. To isolate the effect of dual momentum, we incorporate it into A2 without $\textbf{m}_u$ (\textbf{A4}), which shows consistent improvements over both \textbf{A2} and \textbf{A3}. Finally, combining both dual momentum and mask union yields our complete method, \textbf{DoSAPP}, which achieves the best overall performance. These results underline the individual value of each component and their synergistic effect when integrated. Additional ablations on the sparsity threshold $c$, importance of pseudo label selection from both Teacher-Student, and the role of TTL phases are provided in Appendices \ref{sparsity_ablation_section}, \ref{importance_teacher_logits} and~\ref{ablation_on_ttl_phase}.

\section{Discussion and Conclusion}\label{sec:conclusion}
This work explores how test-time data can be leveraged to improve the retention of previous tasks, drawing inspiration from human learning to build more adaptive models. We show that when used effectively, test-time data is a valuable resource. Our method, DoSAPP, enhances CLIP’s zero-shot performance in the challenging class-incremental setting by combining sparse parameter updates with dual-momentum EMA across supervised and unsupervised phases. \tm{Although our experiments focus exclusively on CIL classification, future work could investigate whether DoSAPP's mechanisms generalize to generative or other transformer-based tasks.} 

\bibliography{main}
\bibliographystyle{tmlr}

\newpage
\section{Appendix / supplemental material}
\label{appendix}
\subsection{Derivation for dual momentum}
\label{proof_dual_mom}
In section \ref{sec:method}, the teacher model parameters $\th^T_i$ undergo exponential moving average as 
\begin{equation}\label{eq:ema_dual_appendix}
   \th^{T}_{i+1} = \boldsymbol{p} \th^{T}_i + \boldsymbol{q} \th^{S}_{i+1}
\end{equation}
where $\boldsymbol{p}$ and $\boldsymbol{q}$ denote the smoothing parameters for the teacher and student model, respectively, and can be computed as 

\begin{gather}
\label{eq:P_Q_ema}
     \boldsymbol{p}  = \alpha_{1} \textbf{m} + \beta_{1}, \notag \\
     \boldsymbol{p}  = \alpha_{2} \textbf{m} + \beta_{2}
\end{gather}


where $\alpha_{i}$ and $\beta_{i}$ for $i \in \{1,2\}$ are the coefficients for the affine transformation of the boolean mask vector $\textbf{m}$.

To account for masked parameters, two momentum values $\delta, \gamma$ are introduced for teacher and student models respectively, such that for the teacher model, affine coefficients $\alpha_1,\; \beta_1$ are computed by solving the equations:
\begin{equation}
     \label{eq:wema_t}
 \begin{aligned}
     \alpha_{1} [\textbf{m}_{ij}=1] + \beta_{1} = \gamma \; , \quad \quad \quad \alpha_{1} [\textbf{m}_{ij}=0] + \beta_{1} = \delta
 \end{aligned}
\end{equation}

and $\alpha_2, \beta_2$ are computed by solving the equations 
\begin{equation}
     \label{eq:wema_s}
 \begin{aligned}
     \alpha_{2} [\textbf{m}_{ij}=1] + \beta_{2} = 1-\gamma\; , \quad \quad \quad  \alpha_{2} [\textbf{m}_{ij}=0] + \beta_{2} = 1-\delta
 \end{aligned}
\end{equation}

This gives 
\begin{equation}
     \label{eq:alpha_1}
 \begin{aligned}
     \alpha_{1} = \gamma-\delta ,\quad  \beta_1 = \delta \; , \\
     \alpha_{2} = \delta-\gamma,\quad \beta_2 = 1-\delta
 \end{aligned}
\end{equation} 

This gives 
\begin{equation}
     \label{eq:P_Q_ema2}
 \begin{aligned}
     \boldsymbol{p} = (\gamma-\delta) \textbf{m} + \delta \; , \\
     \boldsymbol{q} =  (\delta - \gamma) \textbf{m} + 1-\delta
 \end{aligned}
\end{equation}

\subsection{Effect of Momentum ($\gamma, \lambda$) on \small Average Accuracy}
\label{eff_momentum}
\autoref{tab1_clip_mom} shows the sensitivity of our method on the choice of momentum values $\lambda, \delta$ in \autoref{tab1_clip_mom}. A high $\delta$ has been chosen to keep the Teacher model stable, as shown in \citep{mt, dinov2, koh2022online}. It can be seen that when $\gamma = \lambda$ (single momentum EMA), the performance significantly drops. DoSAPP is less sensitive to the choice of $\gamma$, but it highly depends on $\lambda$. We can also see that as $\lambda < \gamma$, the performance again drops.
\begin{table}[h!]
\vspace{12pt}
 \caption{\small Effect of Momentum ($\gamma, \lambda$) on \small Average Accuracy (Acc in \% ), Average Forgetting (F.), and First Task Accuracy (FTA.) *{0.9999, 0.8, 0.9} have been used in the main results.}
    \centering
    \scalebox{1}{
    \setlength{\tabcolsep}{6pt}
    \begin{tabular}{cccc}
    \toprule
    \multirow{2}{*}{\textbf{Momentum} ($\gamma$, $\lambda$)} & \multicolumn{3}{c}{\textbf{Aircraft}} \\
    & Acc. ($\uparrow$) & F. ($\downarrow$) &  FTA. ($\uparrow$) \\
    \midrule
    0.9999, 0.9999 & 23.99 & 18.36 & 12.15 \\
    0.5, 0.9 & 38.41 & 13.27 & 37.64  \\ 
    0.7, 0.9 & 37.22 & 13.05 & 37.72 \\
    \midrule
    0.8, 0.9* & \textbf{39.14} & \textbf{12.55} & \textbf{38.13}  \\ 
    \midrule
    0.8, 0.6 & 37.06 & 15.12 & 29.63  \\
    0.8, 0.5 & 32.95 & 13.40 & 26.33  \\
    
    \bottomrule 
    \end{tabular}
    }
   
    \label{tab1_clip_mom}
\end{table}

\subsection{Hyperparameters} 
\label{hparams}
Table \ref{tab:hparams} shows different hyperparameters that have been used for all the experiments using  CLIP backbones. The hyperparameters were selected based on the performance of the first task of the Cars \citep{cars} dataset. All the results have been gathered over experiments running on Nvidia V100 GPU, averaged over 5 random seeds. 

\begin{table}[ht]
\vspace{12pt}
 \caption{Hyper Parameters for all the experiments using the CLIP ViT-B/16 model.}
    \centering
     \scalebox{1}{
    \begin{tabular}{ccc}
    \toprule
       \textbf{Hparams}  & \textbf{CLIP model}  \\ \midrule
        Batch Size &  64  \\
        Optimizer & AdamW  \\
        Learning Rate & $7.5e-6$ \\
        CL Epochs & 10 \\
        Buffer & 0 \\
        TTL batch size &  64\\ 
        Momentum-EMA ($\delta, \gamma, \lambda$) & 0.9999, 0.8, 0.9  \\
        sparsity ($c$) & 0.1 \\
        \bottomrule
    \end{tabular}
    }
    \label{tab:hparams}
\end{table}

\subsection{Limitation}
\label{limitation}
DoSAPP is a robust algorithm which can be potentially applied to any CL technique for unsupervised adaptation of Test Time Data. However, since it utilises the test data, its primary bottleneck becomes the quality of test data especially if its highly skewed. Another limitation is  the increase in the computational budget due to two deployed models: Teacher-Student framework. We address this by leveraging efficient sparse parameter selection method. \tmh{While DoSAPP remains effective under class imbalance \autoref{tab:imbalanced}, extreme skew or partial coverage in the test-time stream may still bias updates toward overrepresented classes. To improve robustness in such settings, several practical mitigations can be incorporated without modifying the core framework. First, confidence-based update gating can prevent updates when both teacher and student produce low-confidence predictions, reducing drift caused by unreliable pseudo-labels. Second, a balanced mask reactivation strategy can periodically reintroduce a small portion of past-task mask entries based on earlier gradient scores, stabilizing parameters corresponding to tasks that are temporarily absent. Third, slower EMA updates (using a $\lambda$ closer to $\delta$ ) can reduce teacher drift when the observed stream is highly imbalanced. Finally, adaptive learning-rate scaling based on prediction entropy can temper the influence of majority-class samples during TTL. These mechanisms are lightweight, compatible with DoSAPP’s sparse-update design, and improve applicability in real-world settings where test-time data is often skewed.}

\subsection{Broader Impact}
\label{broad_impact}
This work proposes a general algorithm DoSAPP which can be implemented in any of the existing continual learning settings. Further it can be also integrated with other CL algorithms. Therefore, the potential negative societal impacts of our method are similar to those of the Continual Learning algorithms. Generally, DoSAPP would greatly improve performance especially in Test Time Learning. However, as with most CL
algorithms, DoSAPP cannot guarantee to take safe and effective actions in all kinds of scenarios.
We advocate that the users of DoSAPP should be aware of the potential consequences and utilize
DoSAPP safely, especially in online environments such as self-driving, robotics, and healthcare.

\subsection{Dependence on quality of test data used for unsupervised learning}
\label{dep_on_test_data}
We want to highlight that the trained model is expected to generalize to the distribution of the test data. We also assume that any quality degradation will be consistent across time steps. For instance, if the data is corrupted with noise, our method would generalize and adapt the model to this corruption as well. To illustrate this, we conducted a small experiment by adding random Gaussian noise (mean = 0, std = 0.1) to different combinations of the test and evaluation suite (referred to as GN in Table~\ref{tab:Noise}). The results are shown below, with average accuracy (Acc.) followed by forgetting (F.). We observe that when corruption is present in the test-time data, the model is still able to leverage these data and improve on clean evaluation data compared to the test-time baseline by a significant margin of $17\%$ (SPU alone). Interestingly, the model adapted to test-time data with Gaussian noise performs better on evaluation data with Gaussian noise than the case when the test-time data is clean. This is evidence of our method's ability to adapt and generalize to the present test-time conditions.

\begin{table}[ht]
\vspace{12pt}
\caption{Performance of DoSAPP with noise added to $D^u$ and $D^e$ for CIFAR100 Data}
    \centering
     \scalebox{1}{
    \begin{tabular}{cccc}
    \toprule
       \textbf{Test Time Data} $ \boldsymbol{(D^u)}$  & \textbf{Evaluation Data} $ \boldsymbol{(D^e)}$ & \textbf{Acc. ($\uparrow$)} & \textbf{F. ($\downarrow$)}  \\ \midrule
        Clean &	Clean &	79.16 &	7.73 \\
        GN & Clean	& 75.67 &	9.93 \\
        Clean &	GN &	69.50 &	12.86 \\
        GN &	GN &	73.42 &	6.86 \\
        \bottomrule
    \end{tabular}
    }
    \label{tab:Noise}
\end{table}

\begin{figure}[t!]
    \centering
    \begin{subfigure}{0.32\linewidth}
        \centering
        \includegraphics[width=\linewidth]{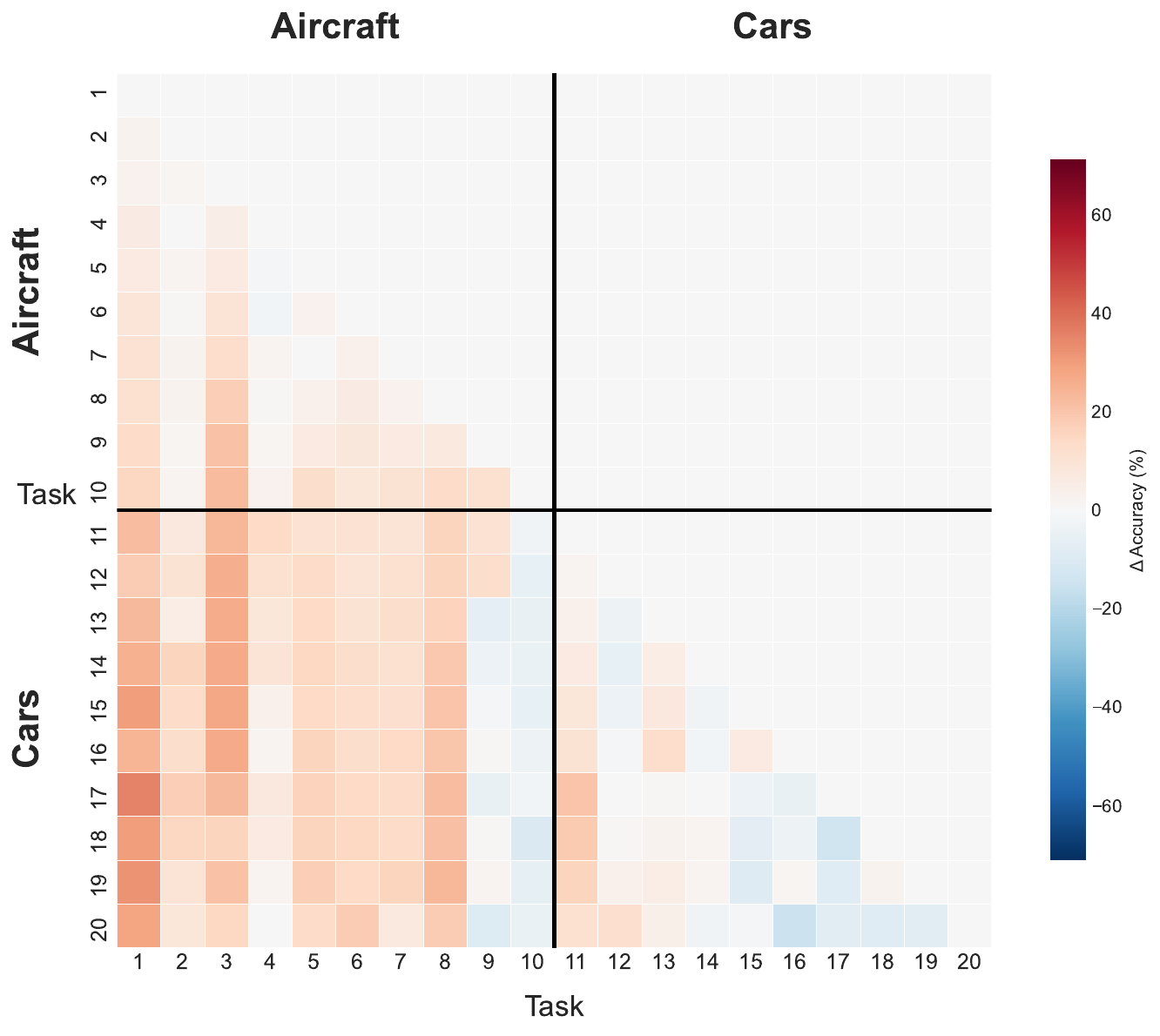}
        \caption{DoSAPP}
    \end{subfigure}
    \hfill
    \begin{subfigure}{0.32\linewidth}
        \centering
        \includegraphics[width=\linewidth]{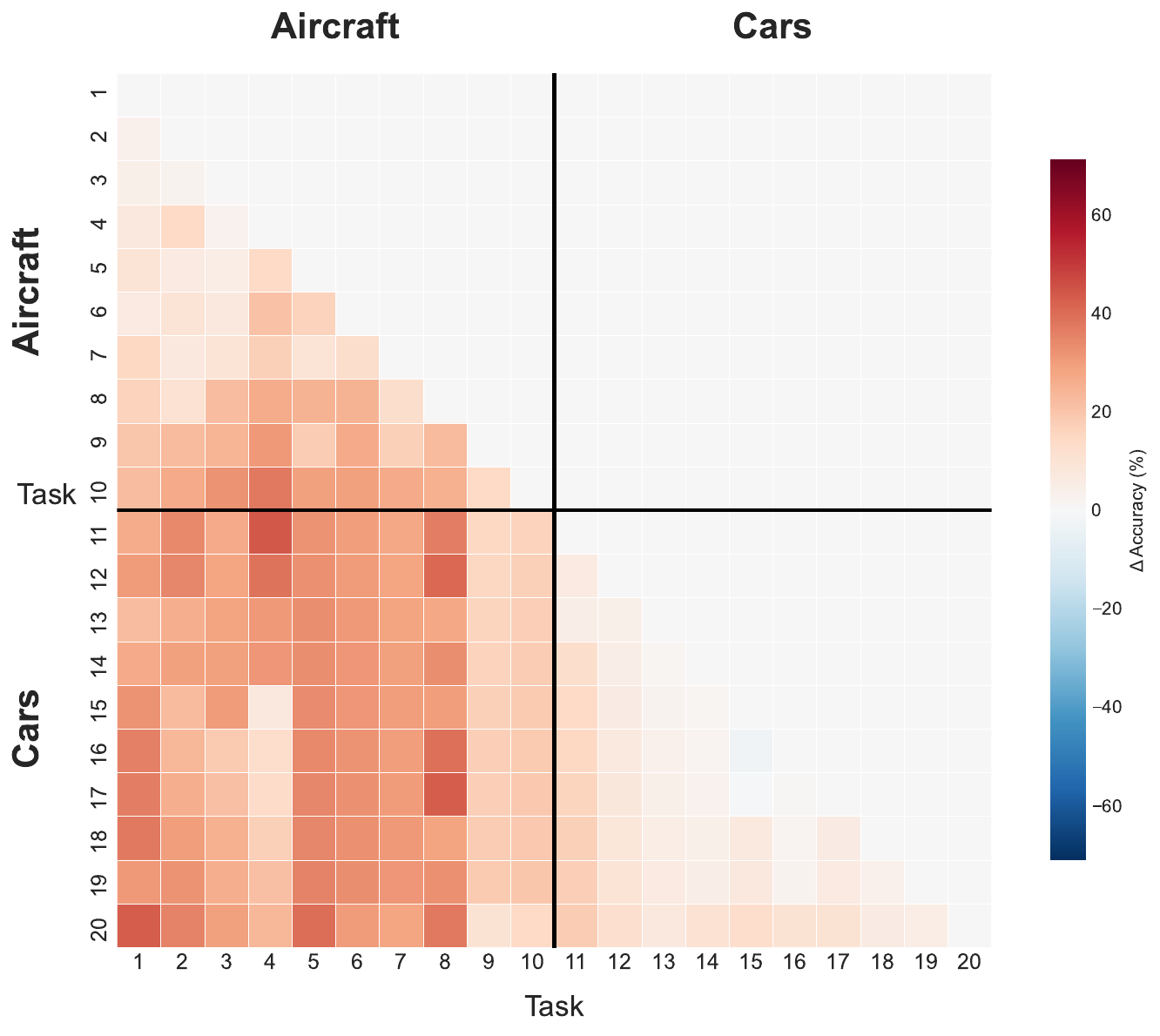}
        \caption{SPU}
    \end{subfigure}
    \hfill
    \begin{subfigure}{0.32\linewidth}
        \centering
        \includegraphics[width=\linewidth]{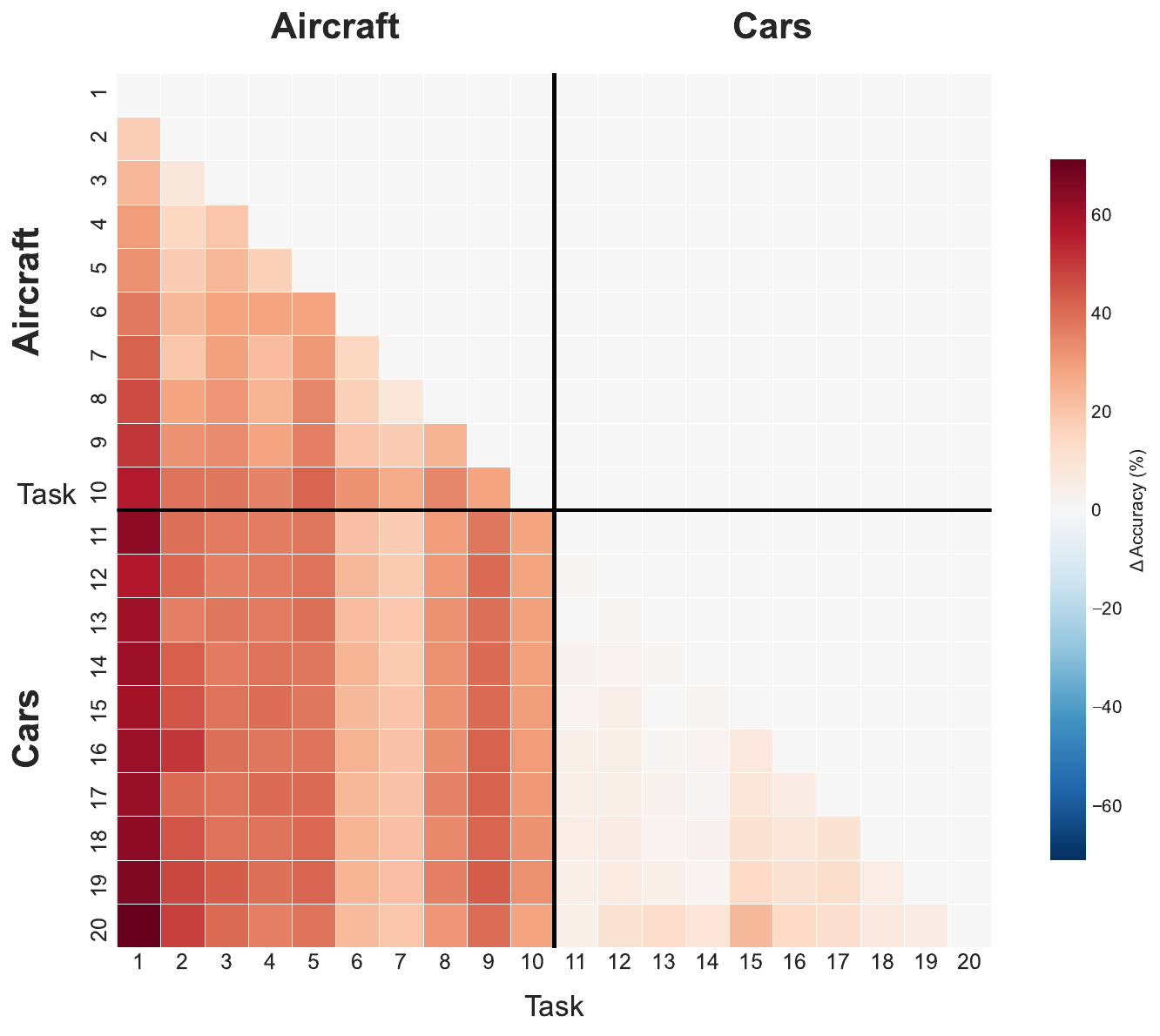}
        \caption{Finetuning}
    \end{subfigure}
    \vspace{1em}
    \caption{
        \textbf{Per-task forgetting matrices} for the long-sequence CIL setting 
        (Aircraft $\rightarrow$ Cars).  
        Each heatmap shows $F_{t,i} = R_{i,i}- R_{t,i}$, i.e., how much performance on task $i$ 
        is lost after learning later tasks.  The vertical and horizontal black lines denote the 
        domain shift from Aircraft (left/top) to Cars (right/bottom).  \textbf{\textit{DoSAPP}} achieves the lowest forgetting across the entire task sequence, with several tasks even exhibiting \textit{negative forgetting}, indicating improved retention as training progresses. In contrast, SPU shows moderate degradation, while standard finetuning undergoes severe catastrophic forgetting in both domains.
    }
    \label{fig:forgetting_heatmaps}
\end{figure}

\subsection{\tmh{Forgetting in Long Sequence scenario with domain shift}} 
\label{appn_forgetting}
To analyze forgetting across the long sequence setting (Aircraft$\rightarrow$Cars),
we compute the \emph{per-task forgetting matrix} $F \in \mathbb{R}^{T\times T}$ from the 
full accuracy trajectory $R$, where $R_{t,i}$ denotes the test accuracy on task $i$ 
after learning task $t$.  
Following standard continual learning analysis, we define the forgetting of task $i$ 
after learning task $t>i$ as
\begin{equation}
    F_{t,i} = R_{i,i} - R_{t,i},
\end{equation}
where $R_{i,i}$ is the accuracy of task $i$ at the moment it is first learned.  
Thus, $F_{t,i}$ quantifies the absolute loss in performance on task $i$ caused by 
subsequent tasks.  We visualize $F$ using heatmaps in Fig.~\ref{fig:forgetting_heatmaps}, 
with a block boundary at the Aircraft--Cars transition.

Across all methods, the domain shift from Aircraft (tasks~1--10) to Cars (tasks~11--20) 
induces substantial forgetting, but the magnitude varies sharply by algorithm.
\textbf{DoSAPP exhibits markedly lower forgetting} throughout both domains,
with only mild degradation in the upper-right block corresponding to 
cross-domain interference. \textbf{SPU shows moderate forgetting}, particularly within 
the Aircraft block, but remains substantially more stable than naive fine-tuning.  
In contrast, \textbf{Finetuning catastrophically forgets} earlier tasks, especially after 
the domain shift, as reflected by large negative values across most of the forgetting matrix.  
These visualizations reinforce the quantitative results in Table~5:
DoSAPP most effectively mitigates long-sequence and cross-domain forgetting.

\subsection{\texorpdfstring{Ablation study about the size of test-time data $D^u$}{Ablation study about the size of test-time data Du}}
\label{size_tta}
In our method, we divided the evaluation data into two halves. One half is for unsupervised learning ($D^u$), and the other half is for evaluation ($D^e$). In the table below, we feed the fraction of $D^u$ for test time learning. $0.25$ means that $25\%$ of the original $D^u$ is fed to the model for unsupervised learning. We notice that when the fraction is below $0.75$, there is an appreciable difference between the performance of our proposed model. However, at $0.75$, the performance is quite close to that of the whole $D^u$. 

\begin{table}[ht]
    \centering
    \vspace{12pt}
     \caption{\small Dependence of the performance of DoSAPP with different proportions of the testing data $D^u$ on the CIFAR100 dataset.}
    \label{tab:dep_on_tta_prop}
     \scalebox{1}{
    \begin{tabular}{ccc}
    \toprule
       \textbf{Fraction of} $\boldsymbol{D^u}$ & \textbf{Acc. ($\uparrow$)} & \textbf{F. ($\downarrow$)}   \\ \midrule
       0.25 & 73.97 & 14.23   \\
        0.5 &  76.83 & 9.44 \\
        0.75 & 79.02 & 8.16 \\
        1 & 79.16 & 7.73  \\
       
        \bottomrule
    \end{tabular}
    }  
\end{table}

\subsection{Ablation on sparsity threshold ($c$)}
\label{sparsity_ablation_section}
We conduct an ablation study on the sparsity threshold ($c$) on different datasets, as shown in \autoref{sparsity_ablation}. We observe that increasing the sparsity threshold (c) beyond 0.1 leads to modest gains, indicating parameter redundancy as the number of updated parameters increases. Conversely, drastically reducing it to 0.01 results in a significant drop in accuracy. This is because the small number of updated parameters do not have enough capacity to accommodate the learned tasks.


\begin{table}[ht]
\centering
\vspace{12pt}
\caption{Effect of Sparsity Threshold ($c$) on Average Accuracy Across Datasets}
\label{tab:sparsity_ablation}
\begin{tabular}{c|ccccc}
\toprule
\textbf{Sparsity ($c$)} & \textbf{Aircraft} & \textbf{Cars} & \textbf{CIFAR100} & \textbf{CUB} & \textbf{GTSRB} \\
\midrule
0.01 & 35.52 & 70.12 & 72.31 & 63.04 & 68.25 \\
0.05 & 37.98 & 72.84 & 77.42 & 65.91 & 70.43 \\
\textbf{0.10} & 39.14 & \textbf{74.87} & 79.16 & \textbf{68.17} & \textbf{72.33} \\
0.30 & 40.48 & 74.12 & \textbf{80.91} & 67.54 & 71.88 \\
0.50 & \textbf{41.49} & 73.65 & 78.44 & 66.82 & 70.02 \\
0.90 & 41.34 & 72.97 & 77.85 & 65.93 & 70.31 \\
\bottomrule
\end{tabular}
\label{sparsity_ablation}
\end{table}

\subsection{Pseudo label only from teacher ($\mathcal{M}_T$) expert}
\label{importance_teacher_logits}
It must be emphasized that pseudo-labeling is not solely dependent on the student model. We further perform an experiment where pseudo labels are given only by the teacher, and it can be observed that the performance across all the datasets deteriorates as shown in the Table \ref{tab:imp_teacher}. \tmh{This is because the teacher model is purposefully updated with very high EMA momentum and adapts slowly to the most recent task. Its logits remain biased toward earlier tasks, leading to outdated pseudo-labels during TTL for most recent tasks at hand. The student, in contrast, is plastic and calibrated for the current task, making its logits critical for accurate pseudo-label selection. Our max-logit expert selection therefore balances teacher stability with student plasticity.}

\begin{table*}[h!]
\vspace{12pt}
\caption{\small Acc. (Average Accuracy, $\uparrow$) and F. (Forgetting, $\downarrow$) comparing DoSAPP with configuration when only Teacher Expert is used in providing pseudo-label at TTL phase.}
    \centering
   
   \resizebox{0.9\textwidth}{!}{
    \setlength{\tabcolsep}{2pt}
    \begin{tabular}{lcc cc cc cc cc}
    \toprule
      \multirow{2}{*}{Description} & \multicolumn{2}{c|}{Aircraft} & \multicolumn{2}{c|}{Cars} & \multicolumn{2}{c|}{CIFAR100} & \multicolumn{2}{c|}{CUB} & \multicolumn{2}{c}{GTSRB} \\
      & Acc. ($\uparrow$) & F. ($\downarrow$) & Acc. ($\uparrow$) & F. ($\downarrow$) & Acc. ($\uparrow$) & F. ($\downarrow$) & Acc. ($\uparrow$) & F. ($\downarrow$) & Acc. ($\uparrow$) & F. ($\downarrow$)\\
    \midrule
     Pseudo-Label$_{\mathcal{M}_T}$   & 36.92 & 13.84 & 71.04 & 0.15         & 75.48 & 8.22 & 65.21 & 2.97 & 68.09 & 1.56 \\   
      \textbf{DoSAPP} & \textbf{39.14} & \textbf{12.55} & \textbf{74.87} & \textbf{-0.74} & \textbf{79.16} & \textbf{7.73} & \textbf{68.17} & \textbf{2.15} & \textbf{72.33} & \textbf{1.02} \\
    \bottomrule
    \end{tabular}
    } 
    \label{tab:imp_teacher}
\end{table*}

\subsection{Ablation on TTL phase:}
\label{ablation_on_ttl_phase}
Below, we provide the ablation showing average accuracy on previously seen tasks (excluding the current task to avoid bias from supervised training) before and after the unsupervised test-time learning (TTL) phase, using the Aircraft dataset. The results show consistent improvement on previous tasks, demonstrating the usefulness of unsupervised TTL as observed in \autoref{ttl_ablation_table}

\begin{table}[ht]
\vspace{12pt}
\caption{\small Average Test Accuracy (referred as Acc. below) Before and After TTL}
\centering
\begin{tabular}{ccc}
\toprule
\textbf{Task} & \textbf{Acc. (Before TTL)} & \textbf{Acc. (After TTL)} \\
\midrule
1  & 78.03 & 78.03 \\
2  & 66.55 & 68.21 \\
3  & 57.32 & 60.13 \\
4  & 50.67 & 54.09 \\
5  & 45.13 & 46.22 \\
6  & 43.39 & 47.58 \\
7  & 39.21 & 45.47 \\
8  & 41.84 & 45.08 \\
9  & 41.97 & 44.14 \\
10 & 37.38 & 39.80 \\
\bottomrule
\end{tabular}
\label{ttl_ablation_table}
\end{table}

We further provide an ablation in \autoref{with_without_ttl} showing average accuracy after training is complete on all the tasks using DoSAPP, with and without TTL phase. We can see that without the TTL phase, the model significantly deteriorates in performance, highlighting the usefulness of test time data.

\begin{table}[ht]
\caption{DoSAPP Ablation Study: With vs Without TTL}

\centering
\begin{tabular}{lcc}
\toprule
\textbf{Dataset} & \textbf{DoSAPP without TTL} & \textbf{DoSAPP with TTL} \\
\midrule
Aircraft & 35.42 & 39.14 \\
CIFAR100 & 67.30 & 79.16 \\
\bottomrule
\end{tabular}
\label{with_without_ttl}
\end{table}

\subsection{Comparing Evaluation on full test set}
Here we compared one of the SOTA CL baselines (SPU \citep{spu}) evaluated on the complete test set, instead of the subset $D_e$. From table \autoref{full_test_bed}, we can observe that DoSAPP when evaluated on the $D_e$, outperforms model trained via SPU method and evaluated on full test set ($D_u \cup D_e$). 

\begin{table*}[ht!]
 \caption{\small Acc. (Average Accuracy, $\uparrow$)  and F. (Forgetting, $\downarrow$) where the latest CL method SPU is evaluated on the complete test set, i.e, without splitting the test set into $D_u$ and $D_e$. DoSAPP is still evaluated on $D_e$ since it utilizes $D_u$ for unsupervised training, and including this subset for evaluation will be unfair. These results show that although the baseline performance of SPU has increased slightly as compared to the performance mentioned in \autoref{results}, DoSAPP outperforms both with and without buffer. }
     \resizebox{\textwidth}{!}{
    \setlength{\tabcolsep}{3pt}
    \begin{tabular}{l cc cc cc cc cc}
    \toprule
     \multirow{2}{*}{\textbf{Method}} & \multicolumn{2}{c|}{\textbf{Aircraft}} & \multicolumn{2}{c|}{\textbf{Cars}}  & \multicolumn{2}{c}{\textbf{CIFAR100}} & \multicolumn{2}{c|}{\textbf{CUB}}  & \multicolumn{2}{c}{\textbf{GTSRB}}  \\
     & Acc. ($\uparrow$) & F. ($\downarrow$) & Acc. ($\uparrow$) & F. ($\downarrow$) & Acc. ($\uparrow$) & F. ($\downarrow$) & Acc. ($\uparrow$) & F. ($\downarrow$) & Acc. ($\uparrow$) & F. ($\downarrow$)\\
    \midrule
    
    SPU & 32.51 & 24.74 & 70.59 & 14.26 & 64.98 & 19.74 & 62.43 & 6.89 & 48.97 & 15.51 \\
    \midrule
    \textbf{DoSAPP} & \bf{39.14} & \bf{12.55} & 
             \bf{74.87} & \bf{-0.74}  &
             \textbf{79.16} & \textbf{7.73} & 
             \bf{68.17} & \bf{2.15} & 
             \bf{72.33} & \bf{1.02} \\
           
           
           
           
    \midrule 
  \rowcolor{grey} SPU + ER=1000 & 44.43 & 14.42  & 77.51 & \textbf{3.26} & 83.99 & -0.39 & 71.51 & 4.84 & \textbf{94.25} & \textbf{-7.87} \\
  
  \midrule

    \rowcolor{grey} \textbf{DoSAPP + ER=200} & \textbf{47.32} & \textbf{8.10} & \textbf{79.17} & 3.92  & \textbf{88.41} & \textbf{-1.96} & \textbf{74.39} & \textbf{2.77} & 83.67 & 1.92 \\


    \bottomrule
    \end{tabular}
    }
    \label{full_test_bed}
\end{table*}

\subsection{Computation time:} 
\label{computation_time}
We include a comparison of wall-clock training time (in seconds) between the full model (all parameters trainable) and the sparse variant (c = 0.1) across 10 tasks and one of the latest SOTA Continual Test Time Adaptation method (RMT) that we have utilized in Table \ref{ttl:comparison} to compare with our proposed algorithm. \tm{As shown in \autoref{training_time}, the sparse variant yields consistent but modest reductions in training time per task. While these improvements are incremental rather than substantial, they demonstrate that restricting updates to a small set of parameters does not introduce additional computational overhead.}

\begin{table}[ht]
\vspace{12pt}
\caption{Training Time Comparison Across Methods}
\centering
\begin{tabular}{cccc}
\toprule
\textbf{Task} & \textbf{All Params} & \textbf{DoSAPP ($\boldsymbol{c=0.1}$)} & \textbf{SPU+RMT} \\
\midrule
1  & 46.61  & 45.69  & 47.13 \\
2  & 70.83  & 68.96  & 71.27 \\
3  & 77.75  & 75.90  & 78.14 \\
4  & 89.55  & 86.92  & 90.07 \\
5  & 101.60 & 98.81  & 102.18 \\
6  & 112.19 & 108.93 & 112.86 \\
7  & 118.28 & 114.71 & 119.03 \\
8  & 130.49 & 127.40 & 131.21 \\
9  & 142.59 & 139.38 & 143.22 \\
10 & 154.01 & 149.65 & 154.77 \\
\bottomrule
\end{tabular}
\label{training_time}
\end{table}

\newpage

\end{document}